# Pragmatist Intelligence
## Where the Principle of Usefulness Can Take ANNs

Antonio Bikić[1], Sayan Mukherjee[1,2]

[1]ScaDS.AI Leipzig, Center for Scalable Data Analytics and Artificial Intelligence, Germany.
[2]Max Planck Institute for Mathematics in the Sciences, Germany.

Corresponding authors: antonio.bikic@uni-leipzig.de, sayan.mukherjee@mis.mpg.de

**Abstract:** Artificial neural networks (ANNs) perform extraordinarily on numerous tasks including classification or prediction, e.g., speech processing and image classification. These new functions are based on a computational model that is enabled to select freely all necessary internal model parameters as long as it eventually delivers the functionality it is supposed to exhibit. Here, we review the connection between the model parameter selection in machine learning (ML) algorithms running on ANNs and the epistemological theory of neopragmatism focusing on the theory's utility and anti-representationalist aspects. To understand the consequences of the model parameter selection of an ANN, we suggest using neopragmatist theories whose implications are well studied. Incidentally, neopragmatism's notion of optimization is also based on utility considerations. This means that applying this approach elegantly reveals the inherent connections between optimization in ML, using a numerical method during the learning phase, and optimization in the ethical theory of consequentialism, where it occurs as a maxim of action. We suggest that these connections originate from the way relevance is calculated in ML systems. This could ultimately reveal a tendency for specific actions in ML systems.

**Keywords:** artificial intelligence, artificial neural networks, pragmatism, neopragmatism, consequentialism, anti-representationalism

## 1. Introduction

The phenomenon of *catastrophic forgetting* (McCloskey et al. 1989) regularly occurs in ANNs when trying to learn tasks sequently. This is an ANN's tendency to entirely "forget" what it has previously learned in order to be able to learn something new during the fine-tuning and adapting state of its learning phase. The problem occurs with data in which the *i.i.d. assumption* (independent and identically distributed random variables) does not hold and poses a problem for sequential learning (Boschini et al. 2022). Numerous approaches have been made to explain and mitigate this phenomenon (Kemker et al. 2017; Kirkpatrick et al. 2017).

Interestingly but so far overlooked is the similarity to a phenomenon occurring in humans: young children, on average, forget a large part of the events happening before the age of three. That means that adults can no longer remember situations and events from this time span. This phenomenon is known as *childhood amnesia* (Bauer 2008, 2015).

Even though both phenomena seem similar, they differ in one particular feature: in humans, only situations and events are forgotten, but not, for example, the skills a person has learned in said situations, nor the knowledge of those persons who remain important in later life, such as parents. There seems, then, to be at least one hypothetical *relevance criterion* by which certain memories are preserved in humans. The question is then: why does it seem permissible for ANNs to eliminate all of its memories?

Catastrophic forgetting can be interpreted as a prime example of a *(neo-)pragmatist approach* (Thayer et al. 1970) put into practice via function approximation. Pragmatism is concerned with useful beliefs. The notion of utility is a pragmatic and parsimonious principle in that it reduces, for example, the components of a system to only those that are necessary for the functioning of the system. Utility also structures risk-taking, i.e., a system takes risks only if profits can be expected. Pragmatist theories of truth (Cormier 2000) incorporate said concept of utility and declare as true any belief or theory that helps getting the results one desires.

Truth, then, is that which is useful. In other words: if it works, it is true. This paradigm is prevalent in engineering and somewhat different from the classical scientific paradigm where theories should correspond with a model of reality.

For the description of the ANN functionality, pragmatism comes in handy because, for one thing, it is free of metaphysics. Second and more importantly, pragmatism offers a decisive advantage: since the actual overall goal (approximation function) (Elbrächter et al. 2021) of an ANN is unknown for an external observer, pragmatic solutions for the achievement of this goal are obvious.

While pragmatism entails some problems as an explanation of living beings for whom it was originally developed, it is perfectly suited to describe the functionality of ANNs. We start with illustrating the key elements of pragmatism in relation to ANNs: 1) a positivist approach to reality, 2) the impossibility to gain knowledge from introspection or intuition, 3) the utility aspect of features and theories that usurp understanding, 4) the denial of representationalism reflected in the abolishment of the idea that the structure of an ANN mirrors reality.

Against this background, we argue that ANNs approximating the overall function becomes the actual goal of the system, rendering human interpretations of the output as side effects. We suggest that it is useless to construct machines capable of understanding human needs when pragmatist approaches realized with function approximation are sufficiently successful without having to construct subjective states as in living beings.

We further propose that ANNs should be viewed in an anti-representationalist manner (Rorty 1979). The way ANNs are constructed excludes the possibility of mirroring the reality inside these networks. Their structures consist of (switching) patterns in search for an optimal set of parameters to fit the target function. These structures need not share any resemblance with a represented view of the world.

We close with an argument on why pragmatism sets the ground for a valid interpretation of ANN-generated functionality in terms of the ethical theory of consequentialism.

## 2. Pragmatist elements in ANNs

In the early connectionist approaches, artificial neurons were invented primarily as a postulated hypothetical species of neurons. Their assembly to ANNs marked a kind of network that is completely described qua computation: ANNs by definition leave out the detailed biology of cells. Furthermore, the fact that the basic level of abstraction is the neuron, also known as the the *neuron doctrine*, is influenced deeply by the useful observation that this level makes a computational level of analysis feasible. So, the chosen level of analysis that provides a framework for the building of ANNs is already useful for computations, since the engineer's template that fosters structure-function relationships plays an important role in construction.

However, there is countervailing empirical evidence suggesting that the neuron does not have to be the only reasonable level to investigate the brain (Chirimuuta 2024, chapter 4).

Understanding the connection between pragmatist approaches and ANNs makes it easier to predict the overall tendencies of ANNs. We start with introducing key elements of pragmatism (Ramberg & Dieleman 2023):

1. a positivist approach to reality
2. the impossibility of gaining knowledge from introspection or intuition
3. the utility aspect of features and theories that usurp understanding
4. the denial of representationalism reflected in the abolishment of the idea that the structure of an ANN mirrors reality

1. *A positivist approach to reality.* ANNs emerge as a prime exemplar of a functionalist (Piccinini, 2004; Chalmers, 1996) and positivist (Park et al. 2019; Friedman 1999) approach to reality, wherein *positivism* is encapsulated by the notion that there is nothing intelligible outside of the realm of measurement. The whole idea of ANNs rests on the *functionalist* premise that everything that can be identified as a function can be constructed by substrate-independently creating an emulation of the original function. For the construction of ANNs, the positivist approach narrows down to a set of features actually exploitable for the achievement of goals based on a measurable reality. In a second step, only those features are selected that are usable for the achievement of a goal, creating a second subset of features. For ANNs, the idea is to create a system that prioritizes model fit over comprehension, prediction and control over explanation and understanding.

2. *Gaining knowledge from introspection or intuition is not possible.* ANNs reflect the line of thought of the behaviorist program (Skinner 1987; Jenkins 1993): relying on measurable aspects of a system, applying the principle of association, and denying a causal impact of subjectively experienced states. ANNs are not geared to be reliant on subjective experience as it is currently not clear how information out of these experiences could be made tangible for the function approximation process. This, in turn, means that every notion of abstract objects or features, i.e., anything not tangible for a positivist approach, are excluded. That said, applying some core behaviorist principles does not mean that ANNs are identical to the behaviorist program, but, considering inter alia methods such as reinforcement learning (Mnih et al. 2013, 2016) or reward shaping (Su et al. 2015), conditioning through positive and negative feedback without having to explain the task (Zhao et al. 2017), and denying the causal impact of subjective states (Schweizer 2009), at least borrows or heavily leans on a decent amount of components from behaviorism.

3. *The utility aspect of features and theories that usurp understanding.* One version of the pragmatist maxim can be stated as follows: true is that which is useful. So, as long as it is not vital for a system to understand, for instance, the sound waves it produces, semantics can be usurped by the pragmatist maxim. Moreover, pragmatism offers a decisive advantage: since the actual overall goal (approximation function) (DeVore et al. 2021) of an ANN is unknown for an external observer, pragmatist solutions for the achievement of this goal are obvious. If an arrangement of hyperparameters, batch sizes, and data sets generates parameters that fit the target function, this arrangement is true as it verifies the statement: the parameterized function works.

4. *The denial of representationalism reflected in the abolishment of the idea that the structure of an ANN mirrors reality.* Trying to understand the inner workings of an ANN, for instance by means of idealization in models (Strevens 2021), is deeply entrenched in the idea of *representationalism* (Papineau 2021), a concept alien to ANNs. Representationalism states that the knowledge generated in a system represents parts of reality (Frigg et al. 2017). Under some conditions, the represented knowledge and objects are exchangeable which means that one can learn something from the representations about that which they represent. As much as this may be true for cerebral cognition, it need not hold for ANNs. In ANNs, the parameters may or may not provide information about the object in the real world, rather ANNs try to fit or approximate the target function by manipulating given parameters. It is therefore unnecessary for these parameters to represent something about the reality as long as they can be simply used to achieve model fit. These parameters are not basic units of meaning like lexemes, instead, they are adjustable elements that are sorted out based on how useful they are for function approximation. If anything, they represent one way to fit the target function but need not tell us anything about the represented object. Since no actual symbols are processed in ANNs, only electronic signals, there is no need to posit that the ANN's structure inherently reflects reality, which amounts to abolishing all common variants of representationalism. Understanding parameters is not a prerequisite for contributing to a functionality that aligns with the loss function.

The predictive powers of ANNs are grounded in the ideas of positivism, functionalism, anti-representationalism, and pragmatism. While the predictive power of ANNs is strong, the question remains, whether there are relevant facts that cannot be funneled through an apparatus of that kind. Since ANNs excel at uncovering correlations in data sets, they compete with hypothesis-driven research. Consequently, aligning oneself overly to these systems carries the risk of developing a form of *epistemic myopia*, leading to a specific form of answers (in form of large regressions).

## 3. Artificial neural networks may narrow our set of theories and knowledge

The success of ANNs is deeply influenced by the common task framework (CTF). CTF involves publicly available data sets, competitors trying to infer a class prediction rule from a data set, and a scoring referee where they can submit their prediction rules. Consequently, the synergy of CTF in combination with predictive modeling directly leads to an overall focus on the optimization of empirical, observable performance (Donoho 2017).

Using most parts of the *intelligence* a system can exhibit to avoid mistakes has always been entrenched in the idea of optimization. This, at first, seems like a winning strategy for any competitive quest where the resources are limited. To understand the predetermined breaking points of such a strategy, it helps to focus on an observation made by the science theorist Michael Strevens (Strevens 2020). Strevens claims that science in general needs a well-dosed and meticulously selected amount of irrationality and that the exclusive approval of empirical evidence is to some extent "irrational". Take aesthetic measures like the *beauty* of a theory: scientists are often guided by the idea of beautiful theories. In physics, for instance, theories containing lots of symmetries are considered to be more beautiful. In most cases, this guidance allows for the improvement or development of theories but it is forbidden to state these principles in scientific journals. As clear as it may have been that the Higgs Boson must exist as it has been predicted, only empirical evidence could verify the prediction resulting from the theory. The same is true for the titanic effort put into the detection of gravitational waves and other comparable undertakings.

Even though we know that aesthetic thought, logical consequence, and the like facilitate progress in science, Strevens claims, that we exclude these principles and focus on the dogma of empirical evidence. There is no need to argue against this dogma as its results clearly justify this approach. However, a relative of this dogma structures the inner workings of ANNs that could influence the predictions of these systems significantly.

Humans know that the occupation with seemingly "useless" disciplines can lead to substantial process in the development of – later to become – disruptive technologies, such as the exploration of prime numbers and their resistance to pattern building that led to modern cryptography algorithms (Shin et al. 2019) and the knot theory that currently helps understanding the protein folding problem (Brems et al. 2022). Both disciplines were considered useless outside of specific mathematical interest for hundreds or even thousands of years. And yet: the evident success of initially "useless" research in various scientific fields has no influence whatsoever on the pragmatist paradigm guiding ANNs.

In order for these parts of mathematics, for example, to be devised further and channeled so that they can be used for real-world problems, one needs theories that can estimate what counts as "right" output. In turn, theories are a product of hypothesis-driven research that is only possible if the set of hypotheses is small. And limiting the set of hypotheses one can make requires, again, a theory. An algorithm finding the best model in the hypothesis space based on a given data set, however, is not based on theory but correlation: we have no good theory to explain comprehensively the functionality.

Function approximation without a guiding theory leads inevitably to results "that just fit". Without the theoretical adjustment of target functions, the pragmatist way of solving problems becomes challenging as pragmatist solutions do not learn about structures deemed useless in achieving model fit. By today's standards, a target function would have to be of paramount complexity to include the development of "useless structures" in the dimensions of a knot theory. It seems, then, that focusing on the development of "useless" structures where the occupation with these structures resembles an end in itself is crucial. Yet, in analogy to the observation Michael Strevens made for science, there seems to be no adjustment in the pragmatist approach for ANNs. We argue that there should be no adjustment in the pragmatist approach seen in ANNs as long as the consequences can be leveraged out.

Being able to tell what one is doing when applying a theory reveals its difference to simply applying a theory ensuring success: the former case relies on using and understanding *rules*. This brings us to the *following of the overall goal*.

## 4. Following the overall goal

When thinking about ANNs, rules guiding the functionality of such networks often structure the notion of these networks. Models with predictive powers are viewed as systems that try to infer class prediction rules from the training data.

Being able to articulate a rule which is followed by a system is not the same as knowing that the system's actions are actually structured by said rule. If a person walking in the cemetery is quiet and one assumes that this person is following some societal rules of reverence, it might also just be the case that the person has a sore throat and would otherwise be speaking, possibly not even knowing about the rule of reverence applicable in this context. Thus, specifying a rule does not come along with the application of that rule.

This intensively discussed problem, known in the current form as the *rule following problem* (Miller et al. 2022; Wittgenstein 2011), is treated in Ludwig Wittgenstein's *Philosophical Investigations*, but has been discussed even earlier in Immanuel Kant's *Critique of Pure Reason*. Though there are plenty of solutions to this problem, it remains still unclear how to objectively determine whether a system is following a rule or just being described by rules.

With the prevalent nominalist view on reality (Field 1980) in science and engineering stating that there are no universals or abstract objects to be found in the empirical world, an ANN cannot necessarily find objective rules or natural imperatives telling it what it "oughts to do". Viewing the same empirical data in another fashion happens only if the theory changes. Since nature cannot make statements of how to improve a theory to make the right predictions, *approximating* the overall goal or the target function becomes the *actual* goal of the system.

An ANN does not have to understand the problem it is facing if it only has to approximate an overall function. As a matter of fact, solving the problem humans interpret into the ANN is one of the many interpretations of the output. Solving human problems becomes a side effect of approximating the overall goal as it is not useful to build machines that focus on the actual problems rather than on an optimization problem.

ANNs have no intended meaning ascribed to the output they generate, as Alan Turing puts it in his seminal work on what was later called the *Turing Test* (Turing 1950). Systems equipped with these strategies function quite well, as the example of the pigeon species *columba livia* (Levenson et al. 2015) capable of "diagnosing" breast cancer shows: these birds have been conditioned to perform such tasks. While having no concept of a diagnosis or illnesses in general, these birds exhibited a behavior (= data) that can be evaluated by external criteria. It does not matter that the birds do not understand the task nor what they think or feel (their subjective states). What matters is the exhibited behavior and its fit to the experimenters' standards.

This makes it clear why the ANNs do not have to actually understand the problems humans make them solve: it would be useless for an ANN to understand the problem since it is not paying attention to the meaning of the input data. Rather, it tries to figure out if the input data can be mapped to the output data. This process does not require machines enabled to process semantic content.

Systems learning with positive and negative reinforcement tend to approach solutions that solve the problem but choose every solution that is not directly forbidden. As it is known from symbolic AI approaches, the explicit formulation of all the rules a system needs to classify, say, a visual object, is not practical. Knowing what is relevant for a situation so that one can solve it traces back to the *frame problem* (McCarthy & Hayes 1969; Dennett 2006). Traditionally, the frame problem was formulated for symbolic AI approaches that represent knowledge. Since ANNs are not representing knowledge and should be described with anti-representationalist approaches, not all the consequences of the frame problem apply. The problem of finding out what data is relevant and should be used to make a decision, however, is closely related to the problem of common sense reasoning, also discussed in the context of the frame problem.

Since humans made ANNs, it is quite clear that they can be interpreted best via function approximation. This is due to the fact that we used this model to build these machines which is why this model could be considered one of the best ways to describe the functioning of these systems. In this regard, pragmatism very clearly becomes the favorable option to operate and understand these machines: the overall function is unknown for the external observer and the machine itself can only channel electronic signals through possible ways, as the positivist account sets the context for action. Humans do not know the function the ANN tries to approximate. ANNs, in turn, only consider and process features that are useful for achieving the given goal. This ultimately leads to the situation where every solution that works must be "right".

When looking at the hardware implementation of ANNs, the values true/false or right/wrong are not measuring characteristics. It is the external idea or objective of proper functionality validating an internal switching pattern. Consistently, these machines are dispensed from any attribution of right/wrong or true/false values except if they confirm a target function. This is in line with the pragmatist approach where the logical value of beliefs is irrelevant as long as they lead to desired outcomes (like putting the clock forward to avoid being late grounds on a false belief).

The utility fixation allows for an enhanced characterization of these systems. As the nine dots puzzle in psychology (Öllinger et al. 2013) reveals, test subjects bring their own innate or acquired constraints to the mathematical puzzle making it hard or impossible to solve. These constraints, however, are not inherent to the mathematical puzzle. The same is true for ANNs: these networks measure model parameters exclusively against utility aspects. The solution set is therefore restricted to a set of possibilities based on the restrictions of utility. This constraint may lead to said limitations when avoiding "useless" structures.

This means that these constraints are actually not structured by rules, but only described in terms of rules. It is helpful for human understanding to imply or assume rules that ANNs obey. As the pigeon example showed, however, the rules we might assume are irrelevant for the systems' successes. The goal of these systems is either completely different or does not exist as a *represented object* inside the system.

## 5. The consequences of pragmatism

In representationalist approaches (Egan 2012), *truth and reference* are fundamental concepts: we cannot conceive of an objective reality that we consider to be real but we can refer to this reality by means of our representations of this reality containing things, or aspects thereof, "as they are". The data we get from this reality is the basis for these representations.

ANNs, in contrast, do not represent knowledge that mirrors reality. When speaking about ANNs, it is crucial not only to focus on the ANN model, meaning a computational model whose components are nothing more than ink on paper as long as they are not interpreted. It is also crucial referring to the concrete hardware implementation. Both instances of ANNs do not need to contain represented knowledge, truth or reference to function properly.

ANNs mark a shift from the view that systems need to constitute their aboutness (what a system is about when it is functioning, e.g. classifying, doing arithmetic, talking etc.) through *reference and truth*. Instead, these systems validate their presumed aboutness through patterns of use. This way of viewing the aboutness or *intentionality* (Lyons 1997; Kriegel 2013) of a system traces back to the behaviorist program: what a system does is what its activity (behavior/functionality) shows.

In an anti-representationalist view (Rorty 1979; Godfrey-Smith 2019), *coping with the world* is a fundamental concept. The goal of anti-representationalism, however, is not seeing the world "as it is". Consequently, anti-representationalism forfeits the distinction between objects as they are and as ANNs make sense of them. There is no underlying knowledge to which a system can refer except the data it gets, thus no actual reference to this kind of underlying knowledge is possible – there is only measured reality. As we showed before, positivism is a fundamental structure of the program of pragmatism, which makes reference and truth inexpedient as the foundational components of such a system.

In an anti-representationalist view, the system and its surroundings are continuous, there is no subject-object separation. Incoming data is transposed into patterns or "habits" that can only fail or persist in the interaction with the empirical world: true are those patterns that prove to be useful in *coping with the world*. These transpositions have nothing to do with the world "as it is" and do not refer to a reality outside the measurement. ANNs do not deal with a world they measure while simultaneously assuming there is a world outside their measurement that is "actually real" and their view is a representation of this real world. Quite the contrary: what is measured *is* what is real. There are no longer necessary and contingent facts, meaning that the structures constituting facts and constituted facts are inseparable, due to the fact that they are not seen as representations and are therefore undistinguishable.

The anti-representationalist component of pragmatism illustrates that the idea of a guiding set of beliefs and theories useful to achieve goals can be applied *mutatis mutandis* to ANNs only. Since beliefs and theories are representations, what is actually chosen is an effective set of parameters that can guide the electronic signals in a way that fits the target function.

As the parameters do not depict or mirror reality in an understandable fashion, they do not inform us about the structures in reality but about successful switching patterns. That said, a relationship between a system and the reality, that is not grounded in representing real-world facts, narrows the perceived reality down to just the parameters that benefit the long- or short-term goals of the target function. In turn, every underlying pattern of a phenomenon that is not useful for the target function is dismissed.

Finding patterns and correlations is crucial for acceptable predictions. However, most theories are thought of as representing the world while omitting a controlled set of properties – scientific theories *idealize* (Strevens 2019). ANNs in contrast focus on parameters encoded in the electronic currents and voltages of, most often, transistors. In ANNs, no structure would be preoccupied with the distinguishing aspects of properties exhibited by objects, since aspects, properties, and objects are of representationalist descent. In that regard, crucial for our argument is the following: if representationalist objects have no consequences in the inner workings of an ANN, the only source of change are useful and useless (switching) patterns. As we showed, these patterns cannot be separated into structures constituting facts and constituted facts. There is no case for creating facts or objects or in general *representations* that are the best approximation to a reality conceivable for ANNs. Rather, patterns need to be found that focus exclusively on the mapping function requiring no further connections to some underlying unreachable reality. This is why it is easier to eliminate useless parameters and patterns, since these structures serve no actual purpose in such a system.

In contrast, representationalist objects like the earlier mentioned "useless" disciplines, e.g., prime number research and knot theory, have a chance of surviving long periods of time until they prove useful. They would, in fact, also survive if never applied in technology.

The task is now to investigate if an anti-representationalist, connectionist system like an ANN needs "useless" structures to achieve the same results as representationalist systems like humans. Or if the development of temporary useless structures hinders ANNs that would otherwise thrive.

## 6. Setting the ground for consequentialist approaches

The pragmatist character of ANNs – thus for most of the machine learning approaches – sets the ground for the common *consequentialist* (Mulgan 2001; Ahlstrom-Vij et al. 2018) interpretation of machine functionality guided by these architectures. Consequentialism assumes that (1) the common good (e.g., well-being) is intrinsically good, (2) the common good can be accounted for by aggregation and (3) that an ethical decision is one with the (relatively) best consequences.

Consequentialism as an ethical theory does not follow from pragmatist approaches, but pragmatist approaches set the stage for consequential theories to be implemented. The transition from model parameter selection to the functionality of a system subordinating its actions to an overall ethical goal hinges on these approaches. In this sense, consequentialist theories align with some core principles of pragmatism:

1. the focus on the utility of actions and the utility of beliefs/theories
2. the context-sensitive approach to situations with no principles that cannot be (at least in theory) violated if for the higher good
3. the dismissal of actions or beliefs that do not benefit the common good or predictions
4. and the view that actions or beliefs are means for the accomplishment of goals.

## 7. Summary

We used the phenomenon of childhood amnesia to show that for an ANN it might be for some reasons permissible to eliminate all of its memories as seen in the phenomenon of catastrophic forgetting. To deepen the analysis of the way ANNs are used to manage their parameter selection, we suggested taking (neo-)pragmatist approaches into account that focus on the concept of usefulness. These kinds of approaches do not rely on, for instance, semantics, to solve a problem.

We suggested that a positivist approach to reality, the denial of gaining knowledge through introspection or subjective states, the focus on utility, and an anti-representationalist interpretation of ANNs are in line with (neo-)pragmatist approaches and describe very well how ANNs work. Due to the anti-representationalist account of these systems, the emphasis is on finding the right switching patterns that fit the target function rather than representing structures of the world. The ANN parameters do not depict reality in an intelligible way, hence they do not inform about structures in reality, but about successful switching patterns when trying to fit a target function.

These systems perform inter alia well because they prioritize model fit over comprehension and prediction over explanation. ANN architectures are not designed to amplify the functionality through making sense of parameters but through exploring parameters that ensure model fit. However, we suggested that this approach, as it favors a specific kind of knowledge, might lead to something like epistemic myopia. Depending on the goals the individual user of an ANN has, this kind of myopia might be a good fit for a specific purpose. However, it could also prevent the pursuit of crucial and alternative ways of knowledge generation leading to insights neopragmatist approaches could not generate.

In this context, we introduced Michael Strevens' idea that science needs a meticulously selected amount of irrationality to function. We applied this insight to ANNs stating that some lines of thought in the history of mathematics seemed "useless" for thousands of years until they became the cornerstones of modern cryptography (prime numbers) and led to the understanding of protein folding (knot theory). These are two vivid examples that at least challenge the idea of strict usefulness leading to a broad understanding of reality. We estimated that (at least by today's standards) a target function would have to be of paramount complexity to include the development of "useless structures" in the dimensions of knot theory, prime numbers, and the like.

Stating that something is useful often shifts the problem by omitting the specification what something is useful *for*. In hypothesis-driven research, good projections can be made if the realm of hypotheses is manageable. But to limit such a set of hypotheses, one needs a theory that exceeds correlation. It would therefore be risky to entrust whole parts of our knowledge generation to ANN-driven machines as this requires to predict what structure can count as "useful". But since this is how ANNs work, we should embrace the questions of *when* to use these systems and *how* to leverage the downsides of a usefulness-driven functioning.

ANNs can omit semantics and representations because they exclusively focus on function approximation where this approximation process *is* the actual goal of these systems making the solving of human problems a by-product. The rules that we assume inside these machines might be irrelevant for the machines themselves, as we showed in the pigeon example.

From an ethical perspective, our findings might offer some insight supporting the interpretation of the ANN's functioning qua consequentialist ethics: even though consequentialism does not follow from (neo-)pragmatism, it could shed light on the transition of model parameter selection to a system that subordinates its actions to an overall (ethical) goal.

# References

Bauer, P. J. "Amnesia, Infantile." In *Encyclopedia of Infant and Early Childhood Development*, edited by Marshall M. Haith and Janette B. Benson, 51-62. San Diego: Academic Press, 2008.

Bauer, Patricia J. "A Complementary Processes Account of the Development of Childhood Amnesia and a Personal Past." *Psychological review* 122 2 (2015): 204-31.

Boschini, Matteo, Pietro Buzzega, Lorenzo Bonicelli, Angelo Porrello, and Simone Calderara. "Continual Semi-Supervised Learning through Contrastive Interpolation Consistency." *Pattern Recognition Letters* 162 (2022): 9-14. https://doi.org/https://doi.org/10.1016/j.patrec.2022.08.006.

Brems, M. A., R. Runkel, T. O. Yeates, and P. Virnau. "Alphafold Predicts the Most Complex Protein Knot and Composite Protein Knots." [In eng]. *Protein Sci* 31, no. 8 (Aug 2022): e4380. https://doi.org/10.1002/pro.4380.

Chalmers, David John. *The Conscious Mind: In Search of a Fundamental Theory.* The Conscious Mind: In Search of a Fundamental Theory. New York, NY, US: Oxford University Press, 1996.

Chirimuuta, M. The Brain Abstracted: Simplification in the History and Philosophy of Neuroscience, MIT Press, 2024.

Cormier, Harvey. The Truth Is What Works: William James, Pragmatism, and the Seed of Death. Rowman & Littlefield Publishers, 2000.

Dennett, D. C. *"Cognitive Wheels: The Frame Problem of AI".* Philosophy of Psychology: Contemporary Readings. New York, NY, US: Routledge/Taylor & Francis Group, 2006.

DeVore, Ronald, Boris Hanin, and Guergana Petrova. "Neural Network Approximation." *Acta Numerica* 30 (2021): 327-444. https://doi.org/10.1017/S0962492921000052.

Donoho, David. "50 Years of Data Science." *Journal of Computational and Graphical Statistics* 26, no. 4 (2017/10/02 2017): 745-66. https://doi.org/10.1080/10618600.2017.1384734. https://doi.org/10.1080/10618600.2017.1384734, especially chapter VI.

Egan, Frances. "Representationalism." In *The Oxford Handbook of Philosophy of Cognitive Science*, edited by Eric Margolis, Richard Samuels and Stephen P. Stich, 250–72: Oxford University Press, 2012.

Elbrächter, D., D. Perekrestenko, P. Grohs, and H. Bölcskei. "Deep Neural Network Approximation Theory." *IEEE Transactions on Information Theory* 67, no. 5 (2021): 2581-623. https://doi.org/10.1109/TIT.2021.3062161.

*Epistemic Consequentialism.* Edited by H. Kristoffer Ahlstrom-Vij and Jeffrey Dunn. Oxford University Press, 2018. doi:10.1093/oso/9780198779681.001.0001.

Field, Hartry H. Science without Numbers: A Defence of Nominalism. Vol. 17: Princeton University Press, 1980.

Friedman, Michael. *Reconsidering Logical Positivism.* Cambridge: Cambridge University Press, 1999. doi:DOI: 10.1017/CBO9781139173193.

Frigg, Roman, and James Nguyen. "Models and Representation." In *Springer Handbook of Model-Based Science*, edited by Lorenzo Magnani and Tommaso Bertolotti, 49-102, 2017.

Godfrey-Smith, Peter. "Dewey and Anti-Representationalism." In *The Oxford Handbook of Dewey*, edited by Steven Fesmire, 151–72: Oxford University Press, 2019.

Jenkins, James J. "What Counts as "Behavior"?". *The Journal of Mind and Behavior* 14, no. 4 (1993): 355-64. http://www.jstor.org/stable/43853775.

Kant, I. (1900ff.). *Gesammelte Schriften*. Berlin/Göttingen: Bd. 1–22 Preussische Akademie der Wissenschaften, Bd. 23 Deutsche Akademie der Wissenschaften zu Berlin, ab Bd. 24 Akademie der Wissenschaften zu Göttingen.

Kemker, Ronald, Angelina Abitino, Marc McClure, and Christopher Kanan. "Measuring Catastrophic Forgetting in Neural Networks." *ArXiv* abs/1708.02072 (2017).

Kirkpatrick, J., R. Pascanu, N. Rabinowitz, J. Veness, G. Desjardins, A. A. Rusu, K. Milan*, et al.* "Overcoming Catastrophic Forgetting in Neural Networks." [In eng]. *Proc Natl Acad Sci U S A* 114, no. 13 (Mar 28 2017): 3521-26. https://doi.org/10.1073/pnas.1611835114.

Levenson, R. M., E. A. Krupinski, V. M. Navarro, and E. A. Wasserman. "Pigeons (Columba Livia) as Trainable Observers of Pathology and Radiology Breast Cancer Images." *PLoS One* 10, no. 11 (2015): e0141357. https://doi.org/10.1371/journal.pone.0141357.

*Ludwig Wittgenstein: Philosophische Untersuchungen*, edited by Savigny Eike von, I-5. Berlin: Akademie Verlag, 2011.

Lyons, William. *Approaches to Intentionality*. Oxford University Press, 1997. doi:10.1093/acprof:oso/9780198752226.001.0001.

McCarthy, John, and Patrick Hayes. "Some Philosophical Problems from the Standpoint of Artificial Intelligence." In *Machine Intelligence 4*, edited by B. Meltzer and Donald Michie, 463--502: Edinburgh University Press, 1969.

McCloskey, Michael, and Neal J. Cohen. "Catastrophic Interference in Connectionist Networks: The Sequential Learning Problem." *Psychology of Learning and Motivation* 24 (1989): 109-65.

Miller, Alexander and Olivia Sultanescu, "Rule-Following and Intentionality", *The Stanford Encyclopedia of Philosophy* (Summer 2022 Edition), Edward N. Zalta (ed.), URL = <https://plato.stanford.edu/archives/sum2022/entries/rule-following/>.

Mnih, Volodymyr, Adrià Puigdomènech Badia, Mehdi Mirza, Alex Graves, Timothy P. Lillicrap, Tim Harley, David Silver, and Koray Kavukcuoglu. "Asynchronous Methods for Deep Reinforcement Learning." Paper presented at the International Conference on Machine Learning, 2016.

Mnih, Volodymyr, Koray Kavukcuoglu, David Silver, Alex Graves, Ioannis Antonoglou, Daan Wierstra, and Martin A. Riedmiller. "Playing Atari with Deep Reinforcement Learning." *ArXiv* abs/1312.5602 (2013).

Mulgan, Tim. *The Demands of Consequentialism*. Oxford University Press, 2001. doi:10.1093/oso/9780198250937.001.0001.

Öllinger, Michael, Gary Jones, and Günther Knoblich. "The Dynamics of Search, Impasse, and Representational Change Provide a Coherent Explanation of Difficulty in the Nine-Dot Problem." *Psychological Research* 78 (2013): 266 - 75.

Papineau, David. "Against Representationalism." In *The Metaphysics of Sensory Experience*: Oxford University Press, 2021.

Park, Yoon Soo, Lars Konge, and Anthony R. Artino. "The Positivism Paradigm of Research." *Academic Medicine* 95 (2019): 690 - 94.

*Phenomenal Intentionality*. Edited by Uriah Kriegel. Oxford University Press, 2013. doi:10.1093/acprof:oso/9780199764297.001.0001.

Piccinini, Gualtiero. "Functionalism, Computationalism, and Mental States." *Studies in History and Philosophy of Science Part A* 35, no. 4 (2004): 811-33. https://doi.org/https://doi.org/10.1016/j.shpsa.2004.02.003.

*Pragmatism. The Classic Writings*. Edited by H. S. Thayer and Robert Paul Wolff. Indianapolis: Hackett Publishing Company, 1970.

Price, Huw. "Naturalism without Representationalism." In *Expressivism, Pragmatism and Representationalism*, edited by Huw Price, Michael Williams, Paul Horwich, Robert Brandom and Simon Blackburn, 3-21. Cambridge: Cambridge University Press, 2013.

Ramberg, Bjørn and Susan Dieleman, "Richard Rorty", *The Stanford Encyclopedia of Philosophy* (Fall 2023 Edition), Edward N. Zalta & Uri Nodelman (eds.), URL = <https://plato.stanford.edu/archives/fall2023/entries/rorty/>.

Rorty, Richard. *Philosophy and the Mirror of Nature*. Princeton: Princeton University Press Princeton, 1979.

Rumfitt, Ian, and Robert B. Brandom. "Making It Explicit: Reasoning, Representing, and Discursive Commitment." *The Philosophical Review* 106 (1997): 437.

Schweizer, Paul. "The Elimination of Meaning in Computational Theories of Mind." In *Reduction*, edited by Hieke Alexander and Leitgeb Hannes, 117-34. Berlin, Boston: De Gruyter, 2009.

Shin, Seung-Hyeok, Won Sok Yoo, and Hojong Choi. "Development of Modified Rsa Algorithm Using Fixed Mersenne Prime Numbers for Medical Ultrasound Imaging Instrumentation." *Computer Assisted Surgery* 24, no. sup2 (2019/10/07 2019): 73-78. https://doi.org/10.1080/24699322.2019.1649070. https://doi.org/10.1080/24699322.2019.1649070.

Skinner, B. F. "Laurence D. Smith. Behaviorism and Logical Positivism: A Reassessment of the Alliance." *Journal of the History of the Behavioral Sciences* 23, no. 3 (1987): 206-10. https://doi.org/https://doi.org/10.1002/1520-6696(198707)23:3<206::AID-JHBS2300230303>3.0.CO;2-V.

Strevens, M. "Permissible Idealizations for the Purpose of Prediction." [In eng]. *Stud Hist Philos Sci* 85 (Feb 2021): 92-100. https://doi.org/10.1016/j.shpsa.2020.09.009.

Strevens, Michael. "The Structure of Asymptotic Idealization." *Synthese* 196, no. 5 (2019/05/01 2019): 1713-31. https://doi.org/10.1007/s11229-017-1646-y.

Strevens, Michael. The Knowledge Machine: How Irrationality Created Modern Science. Liveright Publishing Corporation, 2020.

Su, Pei-hao, David Vandyke, Milica Gasic, Nikola Mrksic, Tsung-Hsien Wen, and Steve J. Young. "Reward Shaping with Recurrent Neural Networks for Speeding up on-Line Policy Learning in Spoken Dialogue Systems." *ArXiv* abs/1508.03391 (2015).

Turing, A. M. (1950) Computung Machinery and Intelligence. *Mind* LIX: 433-460

Zhao, Hang, Orazio Gallo, Iuri Frosio, and Jan Kautz. "Loss Functions for Image Restoration with Neural Networks." *IEEE Transactions on Computational Imaging* 3 (2017): 47-57.